\documentclass[letterpaper]{article} 
\usepackage[submission]{aaai23}  
\usepackage{times}  
\usepackage{helvet}  
\usepackage{courier}  
\usepackage[hyphens]{url}  
\usepackage{graphicx} 
\usepackage{natbib}  
\usepackage{caption} 
\usepackage{algorithm}
\usepackage{algorithmic}

\usepackage{newfloat}
\usepackage{listings}
\DeclareCaptionStyle{ruled}{labelfont=normalfont,labelsep=colon,strut=off} 
\floatstyle{ruled}
\newfloat{listing}{tb}{lst}{}
\floatname{listing}{Listing}

\usepackage{url}            
\usepackage{booktabs}       
\usepackage{amsfonts}       
\usepackage{nicefrac}       
\usepackage{subcaption}

\newcommand{\auc}[1]{{\rm AUC}_{\scriptscriptstyle {\rm FPR}<1\%}}

\title{Improving Out-of-Distribution Detection via Epistemic Uncertainty Adversarial Training}
\author {
    Derek Everett,\textsuperscript{\rm 1}
    Andre T. Nguyen, \textsuperscript{\rm 1, 2}
    Luke E. Richards, \textsuperscript{\rm 3}
    Edward Raff \textsuperscript{\rm 1, 3}
}
\affiliations {
    \textsuperscript{\rm 1} Booz Allen Hamilton\\
    \textsuperscript{\rm 2} Jura Bio, Inc.\\
    \textsuperscript{\rm 3} University of Maryland, Baltimore County\\
    everett\_derek@bah.com, an@jurabio.com, lerichards@umbc.edu, raff\_edward@bah.com
}

\usepackage{todonotes}

\usepackage{bibentry}

\begin{document}

\maketitle

\begin{abstract}
The quantification of uncertainty is important for the adoption of machine learning, especially to reject out-of-distribution (OOD) data back to human experts for review. Yet progress has been slow, as a balance must be struck between computational efficiency and the quality of uncertainty estimates. For this reason many use deep ensembles of neural networks or Monte Carlo dropout for reasonable uncertainty estimates at relatively minimal compute and memory. 
Surprisingly, when we focus on the real-world applicable constraint of $\leq 1\%$ false positive rate (FPR),
prior methods fail to reliably detect OOD samples as such. Notably, even Gaussian random noise fails to trigger these popular OOD techniques. 
We help to alleviate this problem by devising a simple adversarial training scheme that incorporates an attack of the epistemic uncertainty predicted by the dropout ensemble. We demonstrate this method improves OOD detection performance on standard data (i.e., not adversarially crafted), and improves $ \auc\ $  from near-random guessing performance to $\geq 0.75$. 
\end{abstract}

%
\section{Introduction}
\label{sec:intro}
%

Probabilistic, or `uncertainty-aware' machine learning methods have several advantages over their point-estimate counterparts. As an example, propagating epistemic (model) uncertainty to predictions can be essential for catching and avoiding potential model failures \cite{ABDAR2021243}, and avoiding overconfidence. Additionally, the propagation of predictive uncertainty enables active learning, experimental design, and optimal decision making \cite{degroot2004optimal}. The adoption of such models has been somewhat limited by the additional computational overhead required for both training (posterior approximation) as well as prediction (via marginalization). However, recent research has developed methodologies which are both computationally tractable and 
provide reasonably good posterior approximations \cite{gal2016dropout, https://doi.org/10.48550/arxiv.1703.04977,Ritter2018ASL, https://doi.org/10.48550/arxiv.1902.02476, https://doi.org/10.48550/arxiv.2002.08791}.

\citet{smith2018} combined two such approximate methods in Bayesian deep learning, Monte Carlo (MC) dropout \cite{gal2016dropout, Gal2016Uncertainty} and Deep Ensembles \cite{lakshminarayanan2017simple}; we will refer to this combination as \emph{Dropout Ensembles}. They demonstrated that dropout ensembles provide principled estimates of epistemic uncertainty for images both near and far from the training data domain, 
and somewhat more principled uncertainty estimates than a single dropout network. 
The authors attributed this success to independently trained networks exploring different modes (local optima) of the objective function, 
while in each network active dropout samples form a variational approximation of the local basin of attraction, a conclusion also stated in  \citet{https://doi.org/10.48550/arxiv.1912.02757}. 
Thus the combination of ensembling and dropout allows a richer distribution over the model uncertainty than offered by either approach in isolation. 
We note that~\citet{https://doi.org/10.48550/arxiv.2002.08791} were similarly motivated in their development of the MultiSWAG approach. 

The total predictive uncertainty of a classifier, which is given by the entropy, can be decomposed into a sum of two terms. The first term, the \emph{aleatoric uncertainty}, arises from genuine randomness or noise in the data. The second term, the \emph{epistemic uncertainty}, arises from uncertainty in the specification of the model, for instance, uncertainty in its parameters. In particular, the epistemic uncertainty for a deep classifier can be approximated by the Monte Carlo estimate of the \emph{mutual information}, the information we gain about model parameters given a new labeled sample. These independent measures of uncertainty are discussed in detail with corresponding formulae in \citet{smith2018}, therefore we do not repeat the discussion here. However, it is vital to restate the importance in clearly distinguishing between the \emph{total uncertainty}, given by the entropy, and it's independent contributions, the \emph{aleatoric} and \emph{epistemic} uncertainties. \citet{smith2018} clearly demonstrated through their experiments that the epistemic uncertainty (given by the mutual information) vastly outperformed the total uncertainty (given by the predictive entropy) for the task of OOD detection.

Following \citet{smith2018} we will also consider the utility of \emph{epistemic uncertainty} for the task of out-of-distribution (OOD) sample detection. Our study will focus on OOD detection for ``garbage'' fake data and OOD samples from natural image datasets. 
Our approach will employ adversarial training to improve \emph{clean} OOD detection, and this is our primary focus in this manuscript. 
But as a secondary test, we will also consider 
OOD samples altered with  
a first order gradient attack. 
Our goal is measuring and improving the accuracy of OOD detection at low false-positive rates (FPR), given an epistemic uncertainty score. 
Real-world systems often require FPR $\leq 1\%$ because systems with high rates of ``false alarms'' are often deemed unacceptable or unreliable by users~\cite{10.1049/sej.1991.0020}. A conservative lower FPR of $\leq 1\%$ is considered an acceptable ``error in a routine operation where care is required''~\cite{kirwan_1994}. 
By contrast, many works focus on the fully-integrated AUC that disproportionately weights $> 1\%$ FPRs, which would not be acceptable in many real-world systems~\cite{Nguyen2021}.
Therefore, the more relevant and discriminating metric is the standardized partial AUC~\cite{mcclish1989analyzing}, which we denote $\auc\ $. 
This satisfies the property that $0 \leq  \auc\   \leq 1$, where $0.5$ is the performance of random guessing. This has already been implemented in scikit-learn.\footnote{\url{https://scikit-learn.org/stable/modules/generated/sklearn.metrics.roc_auc_score.html#r4bb7c4558997-2}} Moreover, because the TPR is a monotonically increasing function of FPR,  $\auc\ \leq {\rm AUC}$.

The novel contributions delivered in this manuscript include:
1) an experimental demonstration that existing uncertainty based OOD detection methods are not effective at $\leq 1\% \rm{FPR}$, often failing to detect clearly different samples and even pure Gaussian random noise,
2) the development and implementation of a novel uncertainty attack and hardening method that improves OOD detection across fake and real OOD samples for all FPR thresholds, and
3) an experimental demonstration that dropout ensembles trained with our novel method have an increased clean accuracy \emph{and} robustness for the task of OOD prediction.
This result is especially significant in contradistinction to the typically observed \emph{trade-off} between clean \emph{label accuracy} and its robustness~\cite{https://doi.org/10.48550/arxiv.1805.12152}.

%
\section{Related Works}
\label{sec:related_work}
%

The need for reliable classification performance at low false positive rates $\leq 1\%$ (and often $\leq 0.1\%$) has been continual for decades, and usually arises when classification errors have asymmetric costs and benefits. Early work looked at devising methods to explicitly improve low FPR performance for spam detection~\cite{learning-at-low-false-positive-rates}, with more recent works using uncertainty estimates to improve the detection of malware~\cite{Nguyen2021} and malware outliers~\cite{nguyen2022out}. Other works have developed special loss functions to optimize while targeting a desired FPR~\cite{pmlr-v54-eban17a}. The need for better uncertainty estimates, and the current inability to meaningfully detect trivial OOD samples, have been noted as risks for medical applications~\cite{doi:10.1073/pnas.1907377117}. Despite the clear need for both uncertainty quantification and low FPR detection of outliers, we are not aware of any prior studies of uncertainty based OOD detection in the $\leq 1\%$ FPR range. Our study focuses the evaluation on this real-world need~\cite{kirwan_1994}, while also improving OOD detection across all thresholds.

\citet{nguyen2022out} also studied the use of MC dropout over a single network as an approximate Bayesian approach, focusing on the utility of the predicted epistemic uncertainty for discriminating between in-distribution (ID) and out-of-distribution (OOD) samples. In that work, all data considered out-of-distribution were clean samples arising from a natural underlying data distribution (whether images or language text). We will examine the performance of dropout ensembles in the same context of OOD detection via model uncertainty. However, in addition to considering performance in distinguishing clean and natural samples, 
we identify the significant failure of current methods to detect pure noise garbage samples. Our method remedies this failure, and shows the expected performance on this data that should have been easily caught by prior methods. 

Existing studies of predictive uncertainty in the context of adversarial attacks have thus far been quite limited, and focused either on 1) developing methods for classifiers which make more robust \emph{label predictions} \cite{ManunzaPagliardini2021}, or 2) using the predictive uncertainty as a detector (alarm) for flagging samples which have been perturbed by a \emph{label accuracy attack} \cite{smith2018}. 
\citet{ManunzaPagliardini2021} derived an attack to perturb samples in the direction which \emph{maximizes} the \emph{total uncertainty} (epistemic $+$ aleatoric), given by the predictive entropy. Additionally, the purpose of the attack and subsequent adversarial training scheme was focused on increasing adversarial robustness in the \emph{label accuracy}, rather than employing the predictive uncertainty as a useful metric in its own right. Conversely, in this manuscript we specifically target the \emph{epistemic uncertainty}, given by the mutual information. This is in part because our purpose is to explore the utility of the epistemic uncertainty \emph{itself} in downstream tasks (specifically in the context of OOD detection). The epistemic uncertainty was shown to be a better quantity than the predictive entropy for discriminating between ID and OOD samples in \cite{smith2018}. 

\citet{https://doi.org/10.48550/arxiv.1812.05720} also studied the problem of Neural Network over-confidence and OOD detection on OOD samples, including noise and adversarial noise images. However, their uncertainty score was defined by the classifier \emph{confidence} rather than the epistemic uncertainty.\footnote{The confidence, the maximum probability score in the predicted categorical distribution, is directly related to the total predictive uncertainty, not the epistemic uncertainty.}
Additionally, the methods they propose require defining a specific reference noise distribution (OOD) during training, while our robust training method uses \emph{only} information from the in-distribution training data, and the model's predictions.

Additionally, \citet{smith2018} focused on employing the epistemic uncertainty to detect samples which had been attacked by a \emph{label accuracy-targeted attack}, rather than an \emph{uncertainty-targeted attack}. While they did study the epistemic uncertainty specifically for the task of OOD detection, they did not explore the implications of an uncertainty-targeted attack, nor the robustness of predictive uncertainty under such an attack. Their motivations and study were focused on detecting samples which an adversary has designed to fool the model into making \emph{inaccurate label predictions}.

%
\section{Methods for Uncertainty Attacks and Hardening}
\label{sec:methods}
%

We find that adversarial training can be used to improve the OOD performance of a model, provided that the adversarial training is modified to encourage better epistemic uncertainty quantification. 
To do so, we first develop a simple attack targeting epistemic uncertainty. The generation of adversarial examples which attack the model's predicted label accuracy have been well studied \cite{szegedy2014intriguing, goodfellow2015explaining, madry2019deep,carlini2019evaluating}. Sufficient for the purposes of this manuscript is a brief explanation
of the Fast Gradient Sign method (FGSM) \cite{goodfellow2015explaining}. 
Specifically, the FGSM attack is a  first-order (in gradients) attack under the constraint given by the $l_{\infty}$ metric, 
\begin{equation}
    x' = x + \epsilon \rm{sign} (\nabla_x \mathcal{L}(x))
\end{equation}
where $\mathcal{L}(x)$ is the model loss function. Typically, in classification problems, the attacked objective $\mathcal{L}$ is chosen to be the cross-entropy loss between the model predictions and ground truth labels. Thus the FSGM attack alters the input by a perturbation bounded by $\epsilon$ in the direction which would minimize the label accuracy. However, the definition of the FGSM attack invites us to consider perturbing samples to optimize a different objective than the label accuracy. In particular, in this manuscript we explore attacking an objective function which quantifies the epistemic (model) uncertainty. 

Our approach for uncertainty attacks and training begins with our definition of the \emph{Uncertainty Fast Gradient Sign method} (UFGSM) attack method: 
\begin{equation}
\label{eqn:ufgsm}
    x' = x - \epsilon \rm{sign} (\nabla_x \mathcal{U}(x)),
\end{equation}
where $\mathcal{U}(x)$ is the \emph{mutual information} (the Jensen-Shannon Divergence) predicted by the model ensemble. 
Thus, we see for positive (negative) values of $\epsilon$, the attack perturbs the input in a direction to minimize (maximize) the predicted epistemic uncertainty. 
The Monte Carlo approximation of the mutual information for an ensemble of $M$ models, over which each model's dropout mask is sampled $S$ times, is given by 
\begin{equation}
    \mathcal{U}(x) \approx H[\bar{p}(y|D, x)] - \frac{1}{SM}\sum_{i=1}^{S}\sum_{j=1}^{M}H[p_j(y|\omega_i, x)],
\end{equation}
\begin{algorithm}[H]
\caption{Uncertainty Adversarial Training}
\label{alg:uat}
\begin{algorithmic} 
\FOR{epoch in epochs} 
\FOR{$x$, $y$ in batches} 
\STATE loss = 0

\FOR{model in ensemble}
\STATE $\hat{y}$ = model($X$) 
\STATE loss += CrossEntropyLoss($\hat{y}$, $y$) / size(ensemble)
\ENDFOR

\STATE $\epsilon \sim U(0, \epsilon_{\rm max})$
\STATE $x' = x - \epsilon \rm{sign} \nabla_{x} \mathcal{U}(x)$
\STATE loss += $\beta (\bar{\Delta}(x, x') - \bar{\epsilon})^2$
\STATE loss.backward()
\STATE optimizer.step()
\ENDFOR
\ENDFOR
\end{algorithmic}
\end{algorithm}

where $H$ denotes the entropy functional, $\bar{p}$  denotes the average categorical probability distribution (averaged over all models and dropout samples), and $p_j(y|\omega_i, x)$ denotes the categorical probability distribution predicted by model $j$ with fixed weights $\omega_i$ (a single pass of the dropout mask)\cite{smith2018}.

We define the network weight training objective (loss) for uncertainty training by a weighted sum:
\begin{equation}
    \mathcal{L}(x, x') = \mathcal{L}_l(x) + \beta \mathcal{L}_u(x, x'),
\end{equation}
where $\mathcal{L}_l(x)$ is the ordinary cross-entropy loss computed on the clean examples, and
\begin{equation}
    \mathcal{L}_u(x, x') = (\bar{\Delta}(x, x') - \bar{\epsilon})^2
\end{equation}
is a term which 
encourages the model to predict with epistemic uncertainty in proportion to the $l_{\infty}$ distance to the in-distribution data manifold.\\
We achieve this by defining a discrepancy between the
predicted mutual information for clean samples $x$ and attacked samples $x'$,
\begin{equation}
    \bar{\Delta}(x,x') \equiv (\mathcal{U}(x') - \mathcal{U}(x))/H_{\rm max},
\end{equation}
where $H_{\rm max} = \log (K)$ is the maximal predictive entropy for $K$-class classification, 
and $\bar{\epsilon} \equiv \epsilon / \epsilon_{\rm max}$.
This process is described by the pseudo-code in algorithm \ref{alg:uat}.

We see that these equations act similarly to a form of epistemic uncertainty calibration. Rather than calibrating the epistemic uncertainty on a validation dataset, we are instead using the regularization to encourage the model to learn that epistemic uncertainty predicted on a new sample $x'$ should scale in proportion to the $l_{\infty}$ distance between the sample and the training data manifold. We note that this requires \emph{only} the in-distribution training data, and makes no reference to any specific out-of-distribution samples. 
The weighting factor was fixed $\beta=5$ for all experiments. 
This encouraged stronger OOD detection at acceptable reductions in clean label accuracy. We observed the UAT ensemble to have clean label accuracy within 2\% of the Ord. baseline for all experiments except for CIFAR10 and ImNet*, for which the UAT ensemble model yielded 5\% and 7\% reductions. 
Our purpose in this manuscript is to build accurate models for OOD detection, rather than accurate label prediction models. 
The typically observed trade-off between clean \emph{label accuracy} and robustness \cite{https://doi.org/10.48550/arxiv.1805.12152} is an outstanding problem and its resolution is outside of the scope of this work. 

%
\section{Experimental Setup}
\label{sec:expt}
%

In this section, we describe several experiments testing the performance of dropout ensembles for OOD detection, and examine the effects of label (FGSM) and our novel uncertainty (UFGSM) adversarial training methods. 
In all experiments, five Convolutional Neural Networks (CNN) of an appropriate architecture are selected to form each ensemble. 
The five CNNs are trained only on data sampled from the \emph{in-distribution} (ID) dataset. 

%
\subsection{Data and Model Architectures}

The datasets on which we perform experiments include MNIST \cite{deng2012mnist}, Fashion MNIST (FMNIST) \cite{https://doi.org/10.48550/arxiv.1708.07747}, 
CIFAR10 \cite{krizhevsky2009learning}, Street View House Numbers (SVHN) \cite{37648}, Adaptiope (Adapt.) \cite{9423412} and a modified version of the ImageNet dataset (ImNet*) \cite{5206848} 
which includes only two-hundred of the one-thousand classes.\footnote{The two-hundred classes included in our ImNet* dataset are the same classes in the Tiny ImageNet dataset\cite{https://doi.org/10.48550/arxiv.1707.08819}, however we have kept the images at their full resolution (cropped to 224x224 pixels) rather than down-sampling images to 64x64 pixels.}  Additionally, in each experiment we test the model on samples from the Pytorch FakeData dataset (Fake), which generates Gaussian noise.
The CNN architecture for each experiment was selected based on the complexity and dimensionality of the training distribution.

 \begin{table}
  \caption{Description of experiments}
  \label{tab:expts}
  \small
  \centering
  \begin{tabular}{llll}
    \toprule
    Architecture & ID & OOD \#1 & OOD \#2 \\
    \midrule
    CNN & MNIST & FMNIST & Fake  \\ 
    SRN18 & FMNIST & MNIST & Fake  \\
    SRN18 & CIFAR10 & SVHN & Fake \\
    SRN18 & SVHN & CIFAR10 & Fake  \\
    RN18 & ImNet* & Adapt. & Fake  \\
    \bottomrule
  \end{tabular}
 \end{table}

For each experiment we employ either 1) a small custom CNN (SCNN) with dropout, 
2) a modified ResNet18 \cite{https://doi.org/10.48550/arxiv.1512.03385} architecture with smaller kernels appropriate for low-resolution images\footnote{\url{https://github.com/kuangliu/pytorch-cifar}} to which we have added dropout layers (SRN18), or 3) the default Pytorch ResNet18 \cite{https://doi.org/10.48550/arxiv.1512.03385} architecture to which we've added a dropout layer before the final classification (linear) layer (RN18). The configurations of each experiment are listed in table \ref{tab:expts}. 

\subsection{Training Methods}

We compare three different methods for training dropout ensembles. In all experiments except for ImNet*, the image tensor values have been rescaled to $[0, 1]$, and every CNN in the ensemble is initialized with random weights.\footnote{All weights are initialized from the default distribution set by Pytorch, which are uniform distributions with widths inversely proportional to the square-root of the effective number of parameters. } For the ImNet* experiment, we have normalized the image tensor values in accordance with the Pytorch requirement for using pretrained models, and initialized each ResNet18 in the ensemble with the pre-trained weights.\footnote{\url{https://pytorch.org/tutorials/beginner/finetuning_torchvision_models_tutorial.html}}
In all experiments the ensembles are trained for ten epochs using the Adam optimizer with a fixed learning rate of $10^{-3}$.  
First we describe the training data augmentations used in all training methods.

\subsubsection{Training Data Augmentations}
For each of the training methods, the same augmentations were used to prevent the dropout ensembles from overfitting to the training data in both predictive label accuracy as well as predictive uncertainty. Augmentations assist the ensembles in better approximating the underlying ID distribution from which the training data are sampled. This is also essential because we explicitly relate the predictive uncertainty to the $l_{\infty}$ distance to the ID data manifold when we employ uncertainty adversarial training. 

When training on the MNIST and SVHN datasets, we include padding and random cropping and random rotations by up to ten degrees. 
For FMNIST and CIFAR10, the augmentations include padding and random cropping, random rotations by up to ten degrees as well as random horizontal flips. 
Finally, when training on ImNet* we include random cropping, random horizontal flips, and Cutout ~\cite{https://doi.org/10.48550/arxiv.1708.04552} with a patch size of 28 pixels. 

\subsubsection{Ordinary Dropout Ensemble Training}
In the first method, which we denote by \emph{Ord.}, five CNNs are independently trained with random seeds and randomly shuffled (clean) training data. Each CNN is trained completely independently, so the total compute cost to train this model scales linearly with the size of the ensemble. Using NVIDIA Tesla V100s, for each 5 CNN ensemble, training occupied about 
$0.2$ gpu-hours for MNIST,  
$1.25$ gpu-hours for FMNIST and CIFAR10, 
$1.75$ gpu-hours for SVHN, and
$14$ gpu-hours for ImNet*.\footnote{ImNet* training was accelerated by the Fast Forward Computer vision library~\cite{leclerc2022ffcv}.}

\subsubsection{Label Adversarial Dropout Ensemble Training}
The second method, which we call \emph{Label Adversarial Training} (LAT), five CNNs are independently trained with random seeds and randomly shuffled training data, and in each epoch each model is independently trained on an equal mixture of clean and adversarial examples given by the FGSM attack \cite{goodfellow2015explaining}. Each adversarial training batch is generated with attack magnitude $\epsilon$ randomly sampled from a uniform distribution, $\epsilon \sim U(0, \epsilon_{\rm max})$ with $\epsilon_{\rm max} = 0.02$. The total compute cost to train this model scales linearly with the size of the ensemble, and in our experiments each CNN took just under twice as long to train as a corresponding Ord. ensemble member. 

\subsubsection{Uncertainty Adversarial Dropout Ensemble Training} 

In the novel training method which we introduce, \emph{Uncertainty Adversarial Training} (UAT), five CNNs are initialized with random seeds, 
and in each epoch the every model is trained on an equal mixture of clean and adversarial examples given by the UFGSM attack. We note that the prediction of the mutual information required for UFSGM attack is a function of all models in the ensemble, and so in this method the five models can not be trained independently. In practice, we loaded all five models onto the same GPU in order that the mutual information calculation did not require copying tensors between devices. 
On the same Tesla V100s, we found the total cost to train each 5 CNN ensemble
$0.1$ gpu-hours for MNIST,  
$2.8$ gpu-hours for FMNIST and CIFAR10, 
$3.75$ gpu-hours for SVHN, and
$39$ gpu-hours for ImNet*.\footnote{We found fitting five ResNet18s and training full size ImNet* images (3x224x224) impossible on a single gpu with a reasonable batch size. We opted to use the Pytorch DataParallel() method to split each batch across four gpus, which incurred additional slowdowns of inter-device copying.} 

While training, each batch of attacked examples are generated via the UFGSM attack defined in eqn. (\ref{eqn:ufgsm}) with attack magnitude $\epsilon$ randomly sampled from a uniform distribution, $\epsilon \sim U(0, \epsilon_{\rm max})$ with $\epsilon_{\rm max} = 0.02$. The loss function and training method are described in 
the methods section.
In both the LAT and UAT methods, by randomly sampling the attack magnitude for each batch, we discourage the model from overfitting to samples at a fixed $l_{\infty}$ distance from the training data manifold. This is related to the problem called \emph{catastrophic forgetting} in \citet{ijcai2018-520}, where they noted that a model trained with adversarial examples at a \emph{fixed} attack strength did not retain robustness against \emph{weaker} attacks. In their case, attack strength was taken to be the number of PGD iterations. Moreover, \citet{ijcai2018-520} demonstrated that including both weak and strong attacks during training ameliorated the issue of catastrophic forgetting. Our method of sampling the attack magnitude during training is inspired by these results. 

\subsection{Testing Methods and Baselines}

For each experiment, the mutual information predicted by each ensemble is employed as a score for predicting whether samples are ID or OOD. The Monte Carlo estimation is given by 10 dropout samples (forward passes) over each of the five CNNs in the ensemble, totaling 50 samples of the approximation of the posterior predictive distribution.\footnote{This approximation would only recover the true posterior predictive distribution in the limit that
all the local optima to which different CNNs have converged have equal likelihood. An explicitly Bayesian approach would require weighting each model in the ensemble according to its likelihood, but defining such weights is also complicated by the nature of the optimizer.}
\footnote{The cost for making predictions (in gpu-hours) is approximately the same for the Ord., LAT and UAT ensembles, and scales by the product of dropout samples and ensemble size. Predictions made by the DE ensemble only scale with the ensemble size.}
Consequently, we calculate the Receiver Operator Characteristic (ROC) curve as well as its area. 
This is calculated on both clean ID samples, clean OOD samples, and attacked OOD samples. 
Attacked OOD samples are crafted via the UFGSM attack (eqn. \ref{eqn:ufgsm}) with attack magnitudes sampled uniformly $\epsilon \sim U(0, \epsilon_{\rm max})$ with $\epsilon_{\rm max} = 0.02$.

For baselines of comparison, we also test other common methods of OOD detection, including Deep Ensembles ~\cite{lakshminarayanan2017simple} and the maximum softmax (confidence) ~\cite{hendrycks2017a}.
The label adversarially trained ensemble (LAT) \emph{without} an active dropout mask during predictions is equivalent to a deep ensemble of five members, therefore we denote this model by DE in the following tables of results.
\citet{hendrycks2017a} found that the distributions of softmax scores between ID and OOD samples differed, with OOD samples typically yielding significantly less confident scores. Moreover they studied the use of this score for OOD detection. 
This model will be denoted by SM (softmax) in the following tables of results.

\section{Results}

We will now review the results of our UAT method, which dominates prior approaches in terms of OOD detection for fake, clean, and adversarial samples. This makes our method a Pareto optimal improvement, and its improvement increases as the desired minimum FPR becomes smaller. 

\subsection{Mutual Information Separation}



\begin{figure}[H]
\centering
\begin{subfigure}{.5\linewidth}
  \centering
  \includegraphics[width=\linewidth]{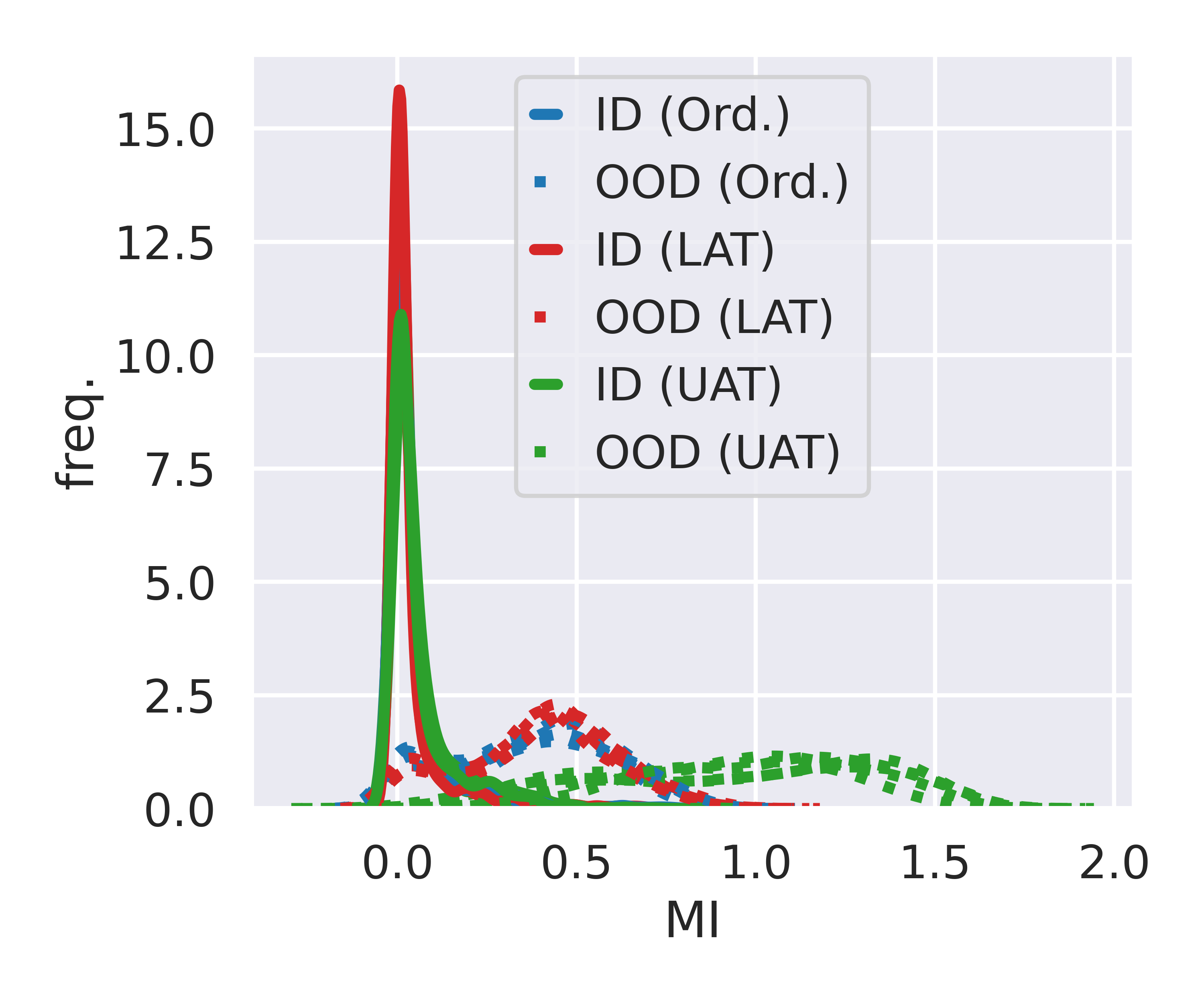}
\end{subfigure}%
\begin{subfigure}{.5\linewidth}
  \centering
  \includegraphics[width=\linewidth]{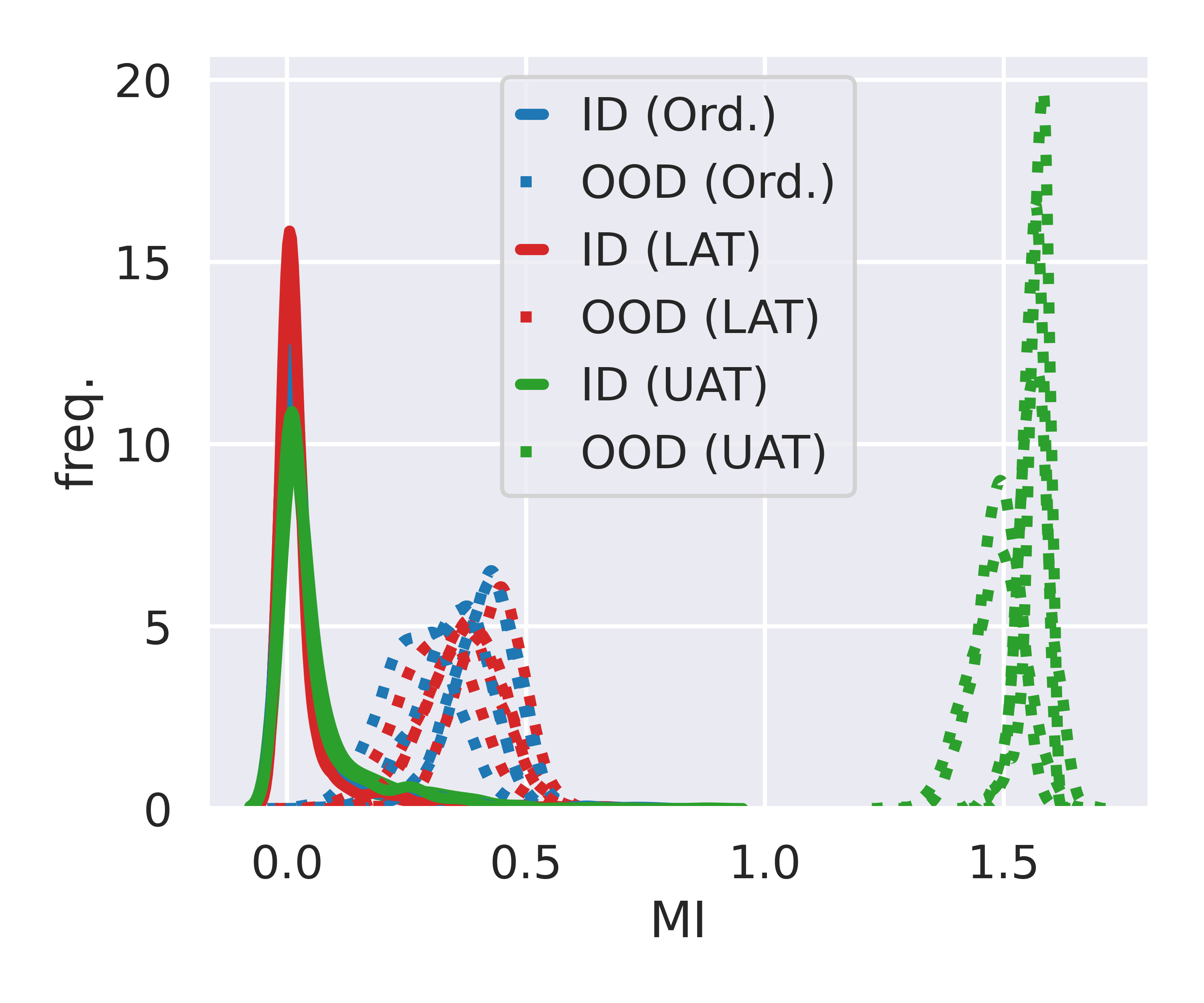}
\end{subfigure}
\caption{Density plot of mutual information (MI) predicted by each ensemble type on both ID (MNIST) and OOD samples, for five independent experiments. OOD samples are from either FMNIST (left) or Fake (right).}
\label{fig:mnist:mi_density}
\end{figure}

In this section we explore the distributions of predicted mutual information on clean ID and OOD samples. In Fig.
\ref{fig:mnist:mi_density} is shown the density of mutual information predicted by each ensemble type which has been trained on MNIST data. Five random experiments (including both training and testing) are shown. In the left subfigure, we see that while all three ensemble types predict similar distributions of mutual information on MNIST (ID) samples, that the UAT ensemble predicts FMNIST (OOD) samples with significantly higher mutual information. 
In the right subfigure we find a similar qualitative ordering, yet with the UAT ensemble predicting FakeData (OOD) samples with a concentrated distribution of large epistemic uncertainty. 

\subsection{OOD Detection on Clean Samples}
%



\begin{figure}[H]
\centering
\begin{subfigure}{.5\linewidth}
  \centering
  \includegraphics[width=\linewidth]{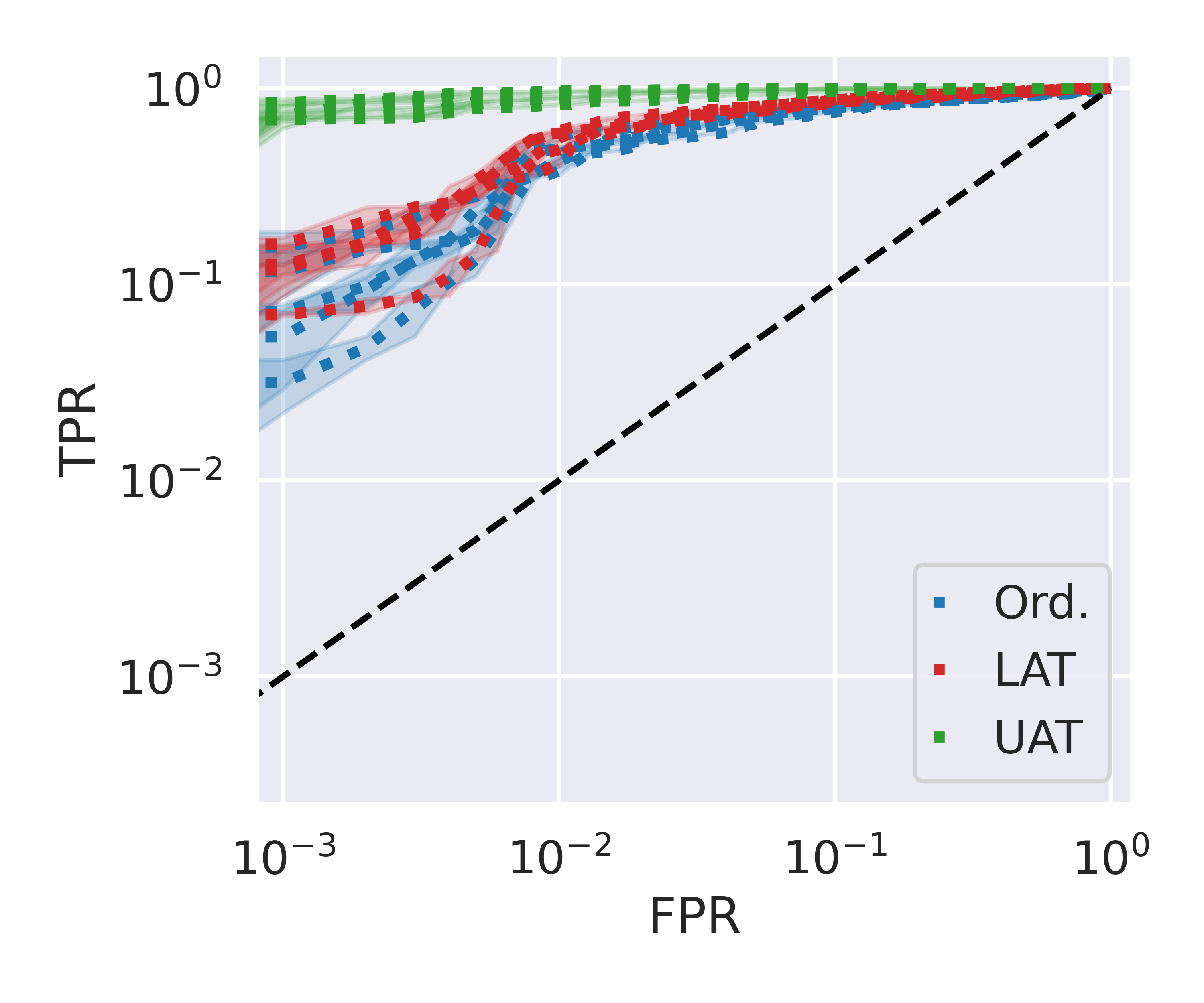}
\end{subfigure}%
\begin{subfigure}{.5\linewidth}
  \centering
  \includegraphics[width=\linewidth]{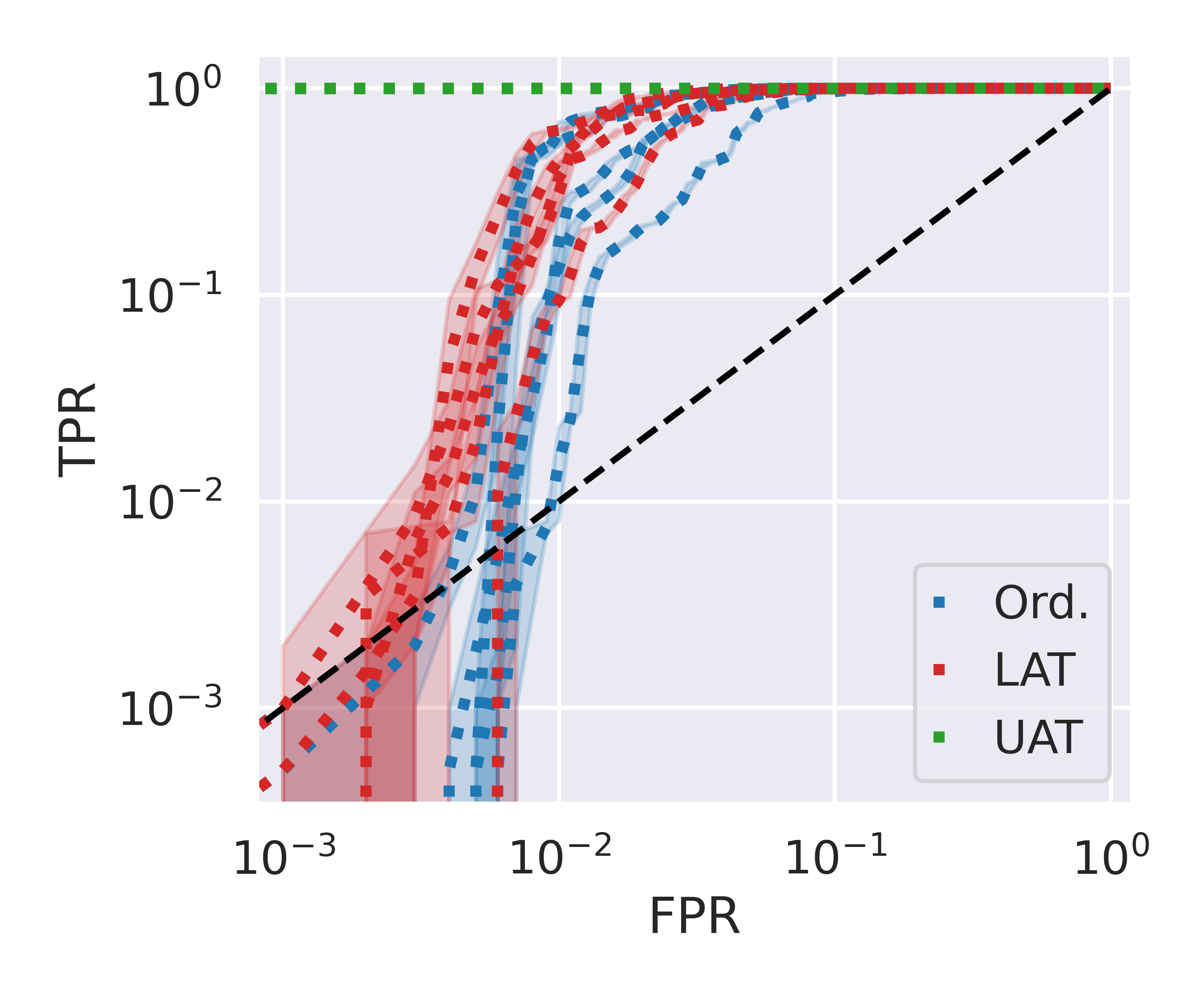}
\end{subfigure}
\caption{ROC curves for distinguishing between clean ID (MNIST) and OOD samples using score given by mutual information for the MNIST experiment. OOD samples are from either FMNIST (left) or Fake (right). Five independent experiments are shown.}
\label{fig:mnist:roc_clean}
\end{figure}

In this section, we plot the Receiver Operator Characteristic (ROC) curves for select experiments, as well as calculate the standardized partial AUC at a maximum false positive rate of 1\% ($ \auc\ $), for clean ID and OOD samples. In Fig.
\ref{fig:mnist:roc_clean} are plotted the ROC curves for the MNIST (ID) trained model in distinguishing between MNIST (ID) test samples and either FMNIST (OOD) samples or FakeData (OOD) samples. 
Focusing on low FPRs illuminates the surprising unreliability of prior methods to distinguish random noise from real data. Interestingly, we see that both Ord. and LAT ensembles are only strong detectors of Fake (garbage) samples down to a FPR of about $10^{-2}$, at which point the TPR drops precipitously to only about $10^{-1}$ (completely unreliable). 
On the other hand, we see that the UAT maintains strong reliability down to FPR of $10^{-3}$.  It is also worth noting that UAT has significantly lower variance in quality of results compared to Ord. and LAT, which improves the reliability of our approach in practice.



\begin{figure}[H]
\centering
\begin{subfigure}{.5\linewidth}
  \centering
  \includegraphics[width=\linewidth]{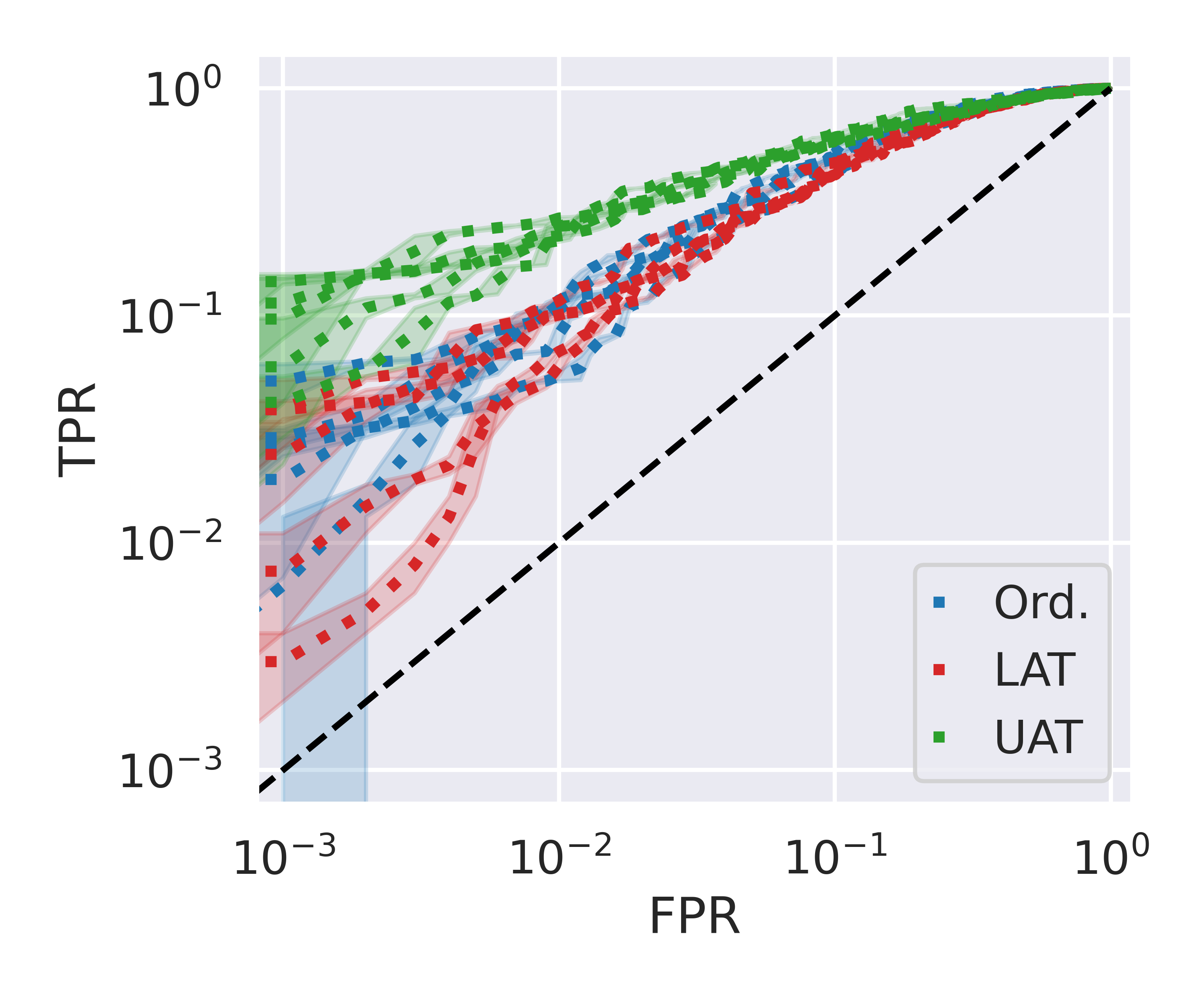}
\end{subfigure}%
\begin{subfigure}{.5\linewidth}
  \centering
  \includegraphics[width=\linewidth]{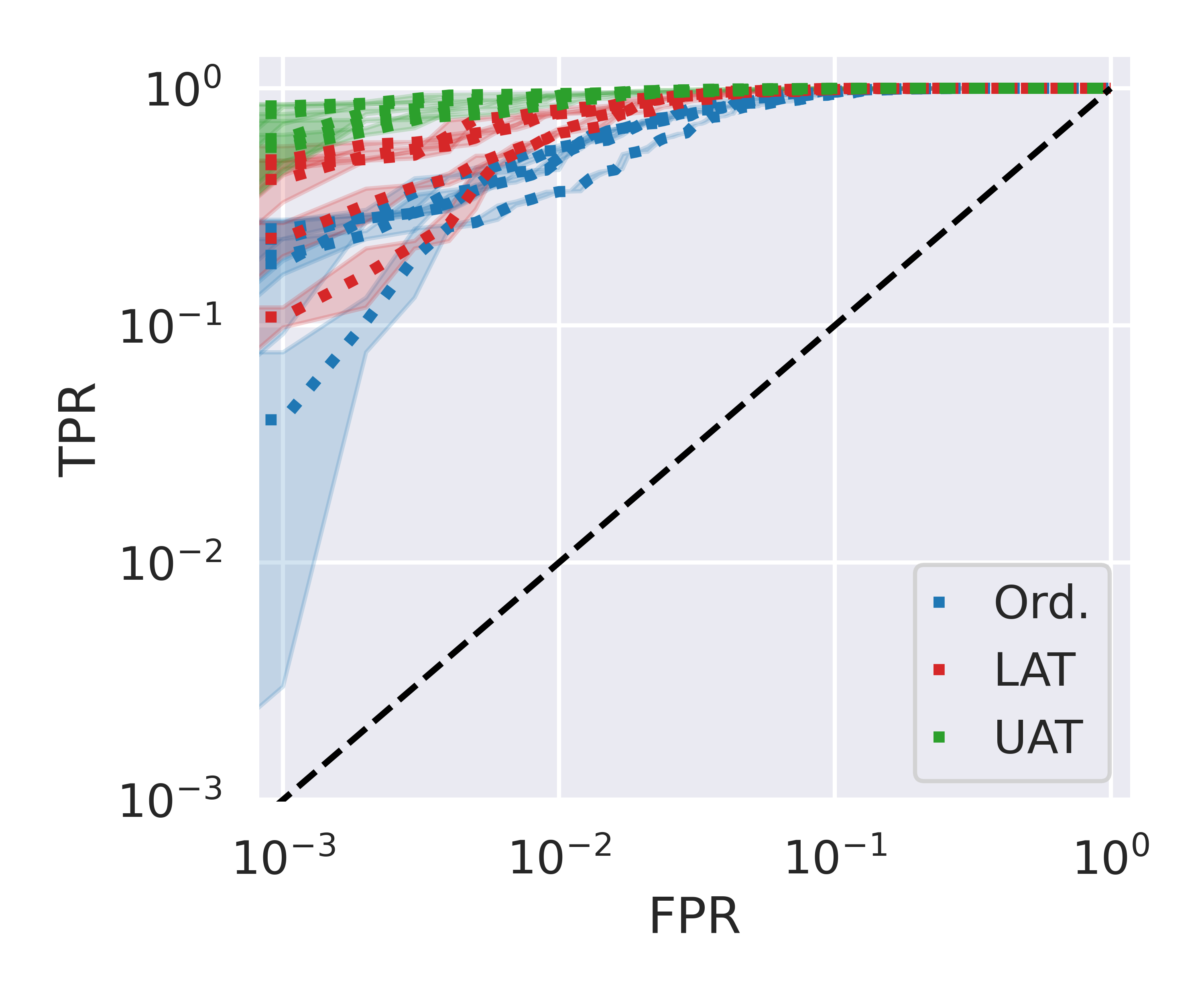}
\end{subfigure}
\caption{ROC curves for distinguishing between clean ID (ImNet*) and OOD samples using score given by mutual information. OOD samples are from either Adapt. (left) or Fake (right). Five independent experiments are shown.}
\label{fig:imnet:roc_clean}
\end{figure}

In Fig.
\ref{fig:imnet:roc_clean} we see the ROC curves for the ImNet* (ID) trained ensembles in distinguishing between ImNet* (ID) samples and either Adaptiope (OOD) samples or FakeData (OOD) samples. 
We see this presents a significantly more challenging task than the MNIST experiment, with each ensemble type having more difficulty.
We still however observe a systematic ordering that UAT ensembles perform better than the rest down to very low FPRs. 
We attribute the difficulty of this task to the inherent difficulty in learning the underlying data distribution of ImNet* data.

\begin{table}[H]
\caption{Standardized partial AUC scores for each ensemble type for \emph{clean} OOD samples.}
\label{tab:auc_clean}
\small
\centering
\begin{tabular}{lllllll}
    \toprule
    & & \multicolumn{5}{c}{$\auc\ $}\\
    \cmidrule{3-7}
    ID & OOD & SM & Ord. & DE & LAT & UAT \\
    \midrule
    MNIST & FMNIST     & 0.53 & 0.62 & 0.55 & 0.65 & 0.92 \\ 
    MNIST & Fake       & 0.50 & 0.53 & 0.50 & 0.55 & 1.00 \\
    FMNIST & MNIST     & 0.68 & 0.97 & 0.58 & 0.99 & 1.00 \\
    FMNIST & Fake      & 0.79 & 0.99 & 0.83 & 1.00 & 1.00 \\
    CIFAR10 & SVHN     & 0.52 & 0.75 & 0.50 & 0.67 & 0.68 \\
    CIFAR10 & Fake     & 0.70 & 1.00 & 0.53 & 1.00 & 1.00 \\
    SVHN & CIFAR10     & 0.51 & 0.75 & 0.51 & 0.71 & 0.97\\
    SVHN & Fake        & 0.51 & 0.89 & 0.50 & 0.87 & 1.00\\
    ImNet* & Adapt. & 0.51 & 0.53 & NA   & 0.52 & 0.58\\
    ImNet* & Fake   & 0.62 & 0.67 & NA   & 0.77 & 0.91\\
    \bottomrule
\end{tabular}
\end{table}

In Table \ref{tab:auc_clean} are listed the $ \auc\ $ scores for all experiments performed. Firstly, we note that in all experiments except for CIFAR10, the UAT ensemble achieves much stronger reliability in recognizing both natural and fake OOD data than the other methods studied. Interestingly, we note that the deep ensemble (DE)~\cite{lakshminarayanan2017simple} method significantly underperforms compared to all other methods studied, showing consistent unreliability in detection of natural and noise OOD samples at low FPRs. We suspect that this is due to the crudeness of its approximate posterior, which collapses all of the probability mass on a handful of points (5 in this work) in the model parameter space. Also, the softmax model (SM) ~\cite{hendrycks2017a} tends to under-perform the Ord., LAT and UAT models. We note that OOD samples tended to yield less confident maximum softmax predictions than ID samples, consistent with the experiments in ~\cite{hendrycks2017a}. However, at the low FPRs studied this quantity becomes less discriminating.

\subsection{OOD Detection on Attacked Samples}
%



\begin{figure}[H]
\centering
\begin{subfigure}{.5\linewidth}
  \centering
  \includegraphics[width=\linewidth]{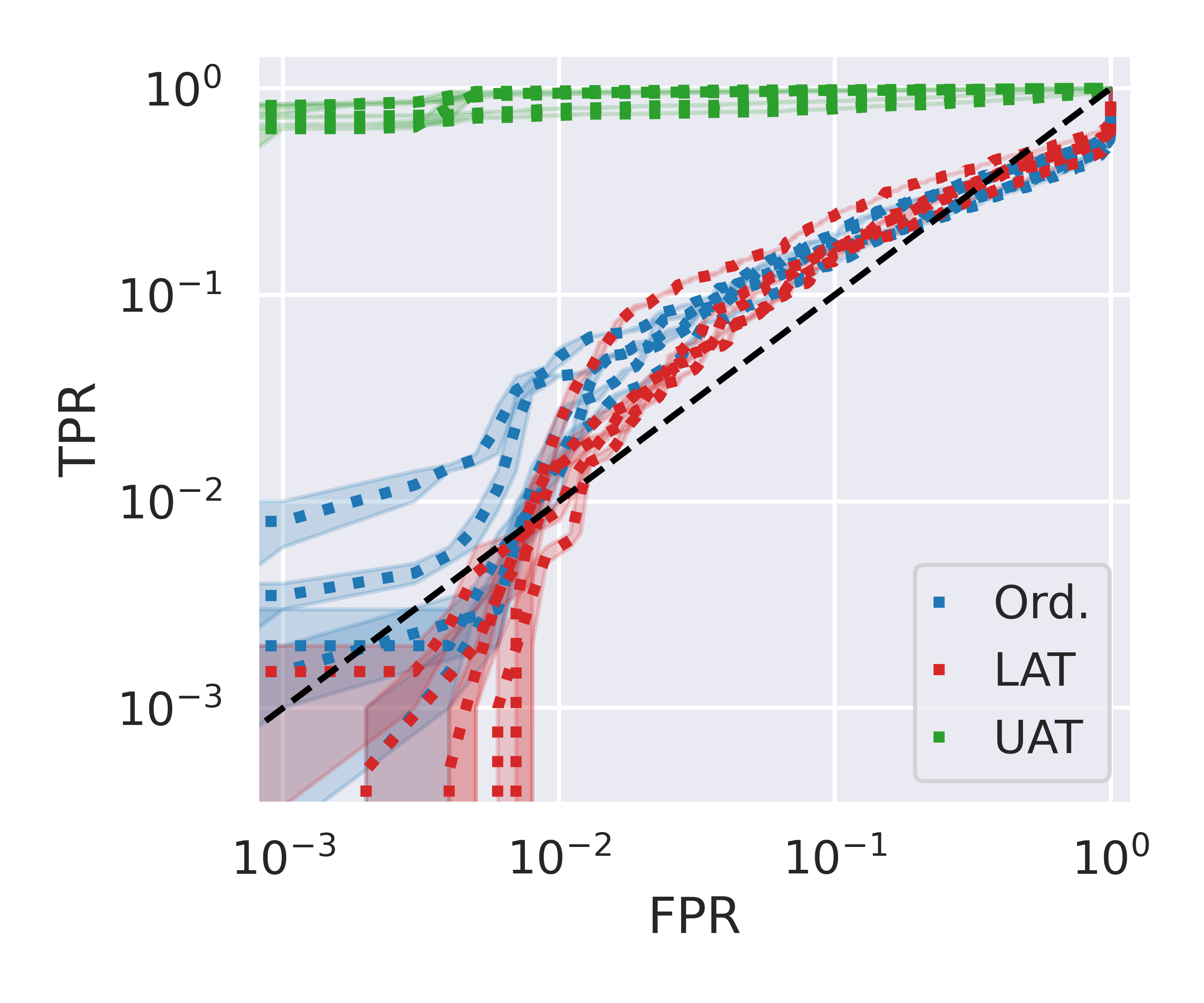}
\end{subfigure}%
\begin{subfigure}{.5\linewidth}
  \centering
  \includegraphics[width=\linewidth]{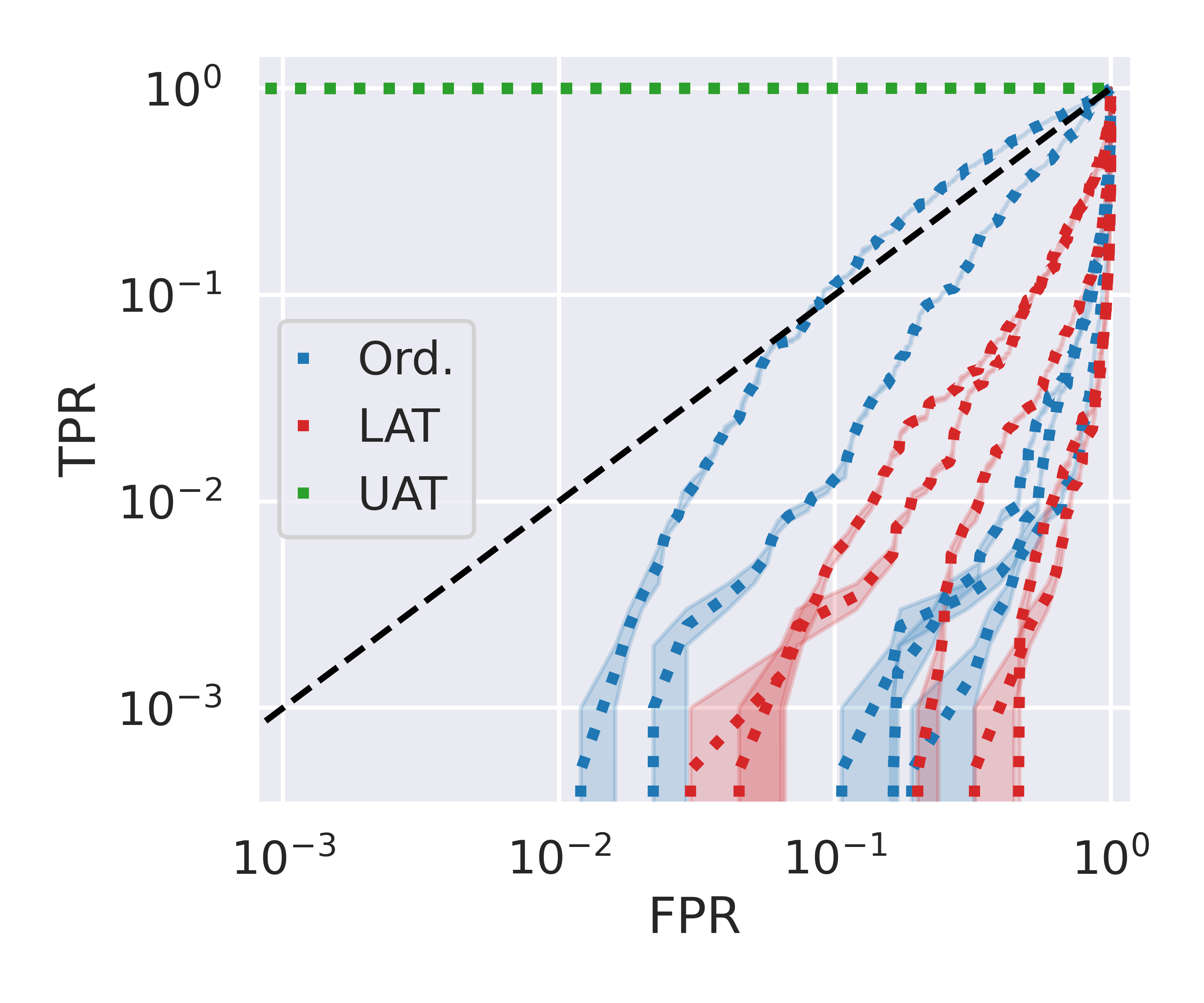}
\end{subfigure}
\caption{ROC curves for distinguishing between clean ID (MNIST) and UFGSM attacked OOD samples using mutual information. Attacked OOD samples are from FMNIST (left) or Fake (right). Five independent experiments are shown.}
\label{fig:mnist:roc_attack}
\end{figure}

In this section, we again study the ROC for select experiments and $ \auc\ $  for all experiments performed, except the task is performed on OOD samples which have been attacked using the UFGSM attack. Because we choose the attack to create a perturbation reducing the epistemic uncertainty, it can be understood to be an attack which would increase the rate of False Negatives (samples which are out-of-distribution but predicted to be in-distribution). 
In Fig.
\ref{fig:mnist:roc_attack} are shown the ROC curves for the MNIST (ID) trained model in distinguishing between MNIST (ID) samples and either attacked FMNIST (OOD) samples or attacked FakeData (OOD) samples. 
We see that both the Ord. and LAT ensembles are very susceptible to the UFGSM attack, 
while the UAT ensemble is (expectedly) very robust, maintaining strong detection performance to FPR of $10^{-3}$. 
This situation is somewhat more extreme when comparing the attacked FakeData examples.



\begin{figure}[H]
\centering
\begin{subfigure}{.5\linewidth}
  \centering
  \includegraphics[width=\linewidth]{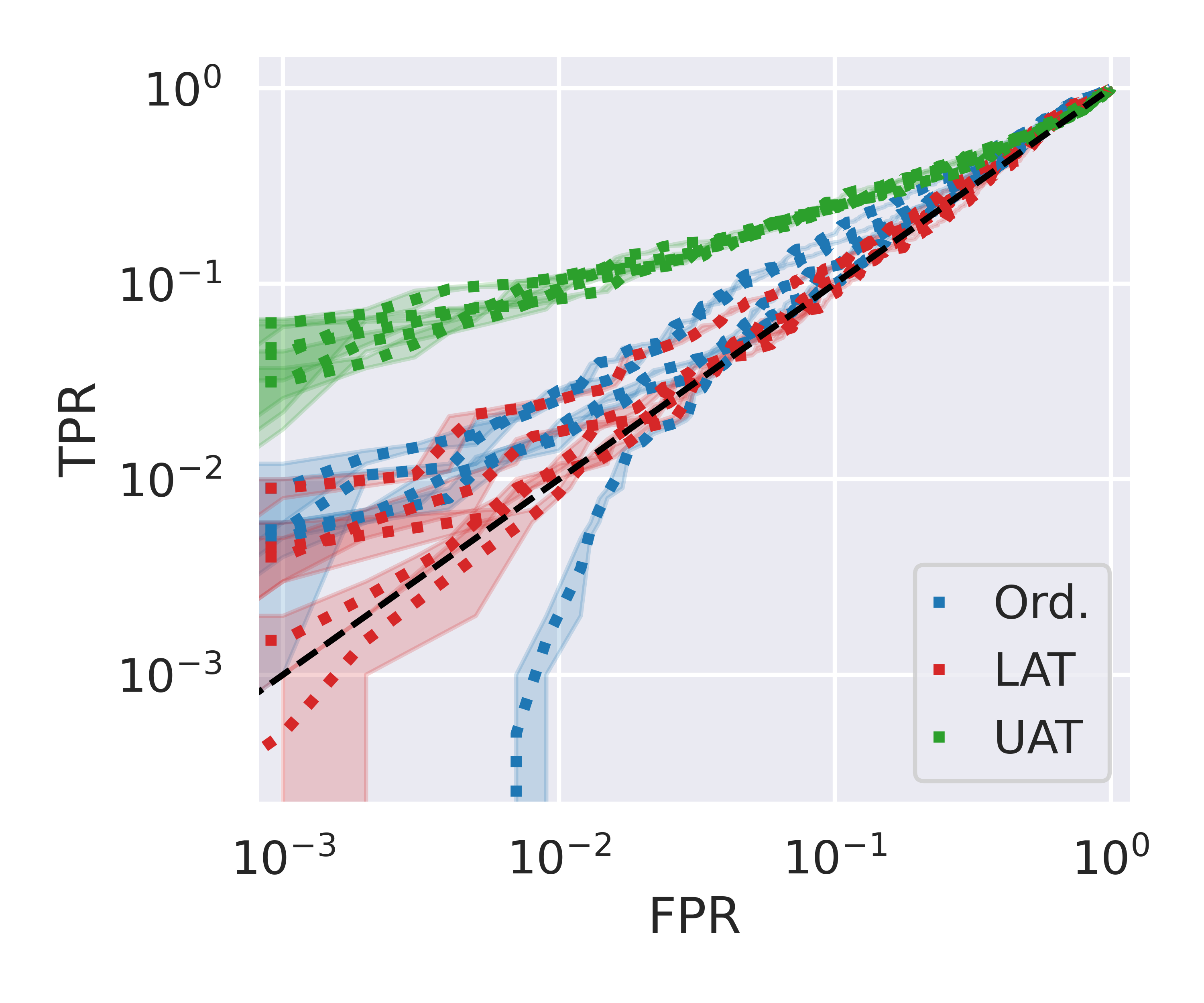}
\end{subfigure}%
\begin{subfigure}{.5\linewidth}
  \centering
  \includegraphics[width=\linewidth]{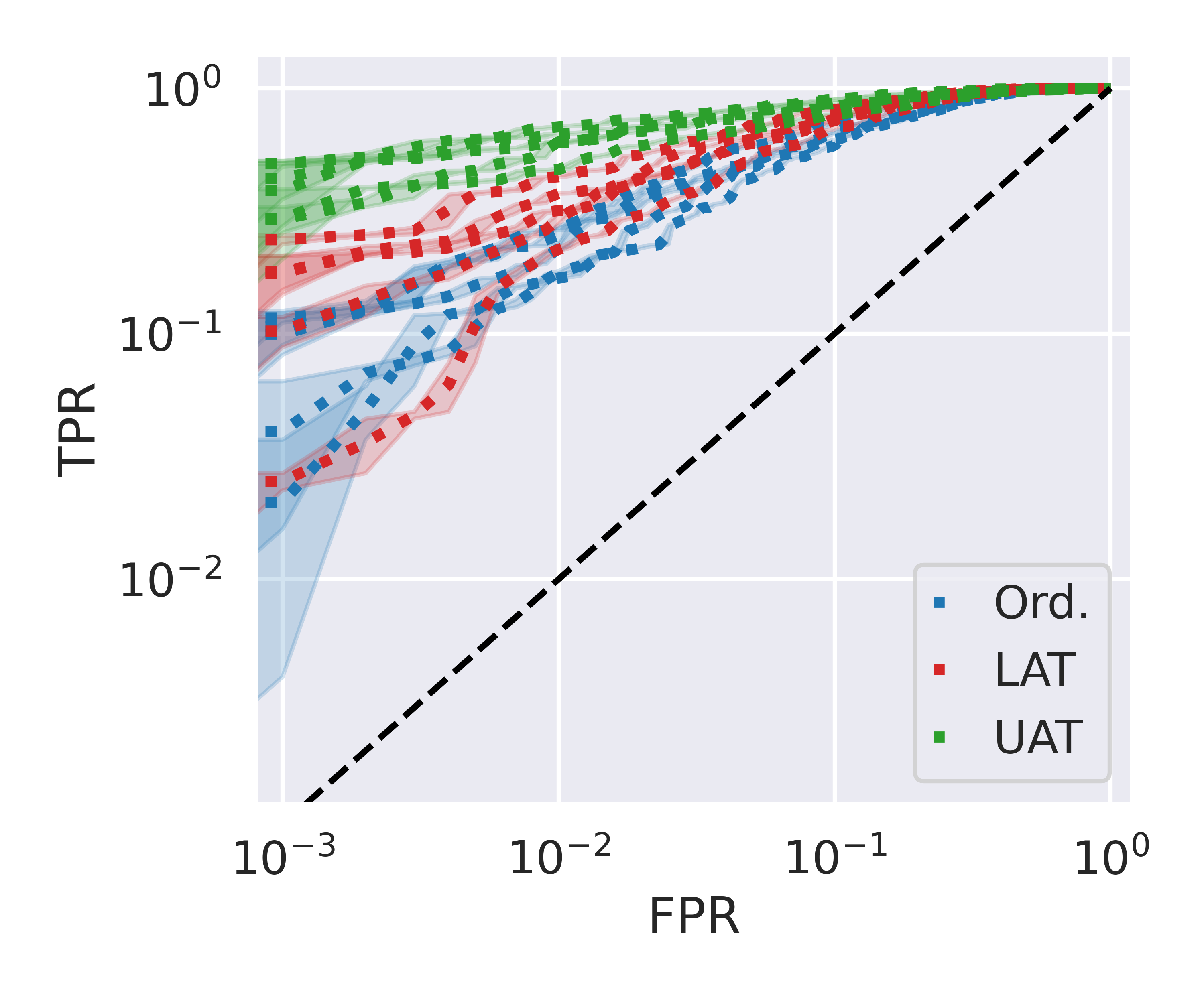}
\end{subfigure}
\caption{ROC curves for distinguishing between clean ID (ImNet*) and UFGSM attacked OOD samples using mutual information. Attacked OOD samples are from Adapt. (left) or Fake (right). Five independent experiments are shown.}
\label{fig:imnet:roc_attack}
\end{figure}

Additionally, in Fig.
\ref{fig:imnet:roc_attack} we see the ROC curves for the ImNet* (ID) trained model in distinguishing between ImNet* (ID) samples and either UFGSM attacked Adaptiope (OOD) samples or UFGSM attacked FakeData (OOD) samples. 
Again, we see that both the Ord. and LAT ensembles are more susceptible to the UFGSM attack, 
while the UAT ensemble (expectedly) shows increased robustness.

\begin{table}[H]
\caption{Standardized partial AUC scores for each ensemble type for UFGSM \emph{attacked} OOD samples. Best results in \textbf{bold}. }
\label{tab:auc_attack}
\small
\centering
\begin{tabular}{llllll}
    \toprule
    & & \multicolumn{4}{c}{$ \auc\ $}\\
    \cmidrule{3-6}
    ID & OOD & Ord. & DE & LAT & UAT \\
    \midrule
    MNIST & FMNIST     & 0.50 & 0.52 & 0.50 & \textbf{0.91} \\ 
    MNIST & Fake       & 0.50 & 0.50 & 0.50 & \textbf{1.00}\\
    FMNIST & MNIST     & 0.69 & 0.57 & 0.64 & \textbf{0.99}\\
    FMNIST & Fake      & 0.70 & 0.73 & 0.67 & \textbf{1.00}\\
    CIFAR10 & SVHN     & 0.57 & 0.50 & 0.50 & \textbf{0.84}\\
    CIFAR10 & Fake     & 0.60 & 0.50 & 0.56 & \textbf{1.00}\\
    SVHN & CIFAR10     & 0.51 & 0.50 & 0.50 & \textbf{0.95}\\
    SVHN & Fake        & 0.55 & 0.50 & 0.51 & \textbf{1.00}\\
    ImNet* & Adapt. & 0.50 & NA   & 0.50 & \textbf{0.53}\\
    ImNet* & Fake   & 0.57 & NA   & 0.61 & \textbf{0.75}\\
    \bottomrule
\end{tabular}
\end{table}

In Table \ref{tab:auc_attack} are listed the $ \auc\ $  scores on UFGSM attacked samples for each ensemble type for each experiment performed.
Noticeably, the Ord., DE and LAT ensembles barely outperform random-guessing at the low FPRs studied, illustrating their complete vulnerability to the attack. 
However, for nearly all experiments the UAT ensemble maintains reliability under attack.

%
\section{Conclusions and outlook}
\label{sec:conclusions}
%

In this manuscript, we have studied the quality of the epistemic uncertainty predicted by dropout ensembles, as well as their robustness to epistemic uncertainty targeting attacks. We have demonstrated via several computer vision experiments that deep ensembles, MC dropout ensembles, and confidence based models  
fail to discriminate in- and out-of-distribution samples at the low false positive rates necessary for deployment, 
Additionally, we have shown that epistemic uncertainty predicted by deep ensembles and MC dropout is very susceptible to epistemic uncertainty targeting attacks. 
Finally, we have demonstrated that Uncertainty Adversarial Training significantly improves both the accuracy in detecting clean OOD samples, and
robustness against epistemic uncertainty targeting attacks. 
Although the uncertainty targeting attack we have employed, UFGSM, is a first-order attack method, this attack can easily be iterated to define stronger attacks~\cite{https://doi.org/10.48550/arxiv.1607.02533}, the study of which we leave for a future work. 

\bibliography{aaai23}

\end{document}